\documentclass{article}
\usepackage{ijcai23}

\usepackage{times}
\usepackage{soul}
\usepackage{url}
\usepackage[hidelinks]{hyperref}
\usepackage[utf8]{inputenc}
\usepackage[small]{caption}
\usepackage{graphicx}
\usepackage{amsmath}
\usepackage{amsthm}
\usepackage{booktabs}
\usepackage{algorithm}
\usepackage{algorithmic}
\usepackage[switch]{lineno}

\usepackage{url}
\usepackage{wrapfig}
\usepackage{graphicx}
\usepackage{multirow}
\usepackage{makecell}
\usepackage{booktabs}
\usepackage{amssymb}
\usepackage{longtable}
\usepackage{xcolor}
\usepackage{colortbl}
\usepackage{hhline}
\usepackage{color}
\usepackage{dsfont}
\usepackage{multirow}
\usepackage{bbding}
\usepackage{pifont}% http://ctan.org/pkg/pifont
\def\eg{\emph{e.g., }}  

\def\ie{\emph{i.e., }}

\def\etal{\emph{et al.} }

\title{IJCAI--23 Formatting Instructions}

\author{
    Author Name
    \affiliations
    Affiliation
    \emails
    email@example.com
}

\usepackage[misc]{ifsym}
\title{Graph Propagation Transformer for Graph Representation Learning}

\author{
Zhe Chen$^{1}$
\and
Hao Tan$^{2}$\and Tao Wang$^{1}$\and Tianrun Shen$^{1}$\and Tong Lu$^{1}$ \\
Qiuying Peng$^{2}$\and Cheng Cheng$^{2}$\And Yue Qi$^{2}$ \\
\affiliations
$^1$State Key Lab for Novel Software Technology, Nanjing Univerisity \\
$^2$OPPO Research Institute \\
\emails
chenzhe98@smail.nju.edu.cn,
lutong@nju.edu.com
}

\begin{document}

\maketitle

\begin{abstract}
This paper presents a novel transformer architecture for graph representation learning. The core insight of our method is to fully consider the information propagation among nodes and edges in a graph when building the attention module in the transformer blocks. Specifically, we propose a new attention mechanism called Graph Propagation Attention (GPA). It explicitly passes the information among nodes and edges in three ways, \ie node-to-node, node-to-edge, and edge-to-node, which is essential for learning graph-structured data. On this basis, we design an effective transformer architecture named Graph Propagation Transformer (GPTrans) to further help learn graph data. We verify the performance of GPTrans in a wide range of graph learning experiments on several benchmark datasets. These results show that our method outperforms many state-of-the-art transformer-based graph models with better performance. The code will be released at \url{https://github.com/czczup/GPTrans}.
\end{abstract}

\section{Introduction}

In many real-world scenarios, information is usually organized by graphs, and graph-structured data can be used in many research areas, including communication networks and molecular property prediction, etc. 
For instance, based on social graphs, lots of algorithms are proposed to classify users into meaningful social groups in the task of social network research, which can produce many useful practical applications such as user search and recommendations. Therefore, graph representation learning has become a hot topic in pattern recognition and machine learning~\cite{cai2020graph,ying2021graphormer,brossard2020graph}.

With the development of deep learning, many methods have been developed for graph representation learning~\cite{perozzi2014deepwalk,zhang2019shne,ying2021graphormer,hussain2021egt,rampavsek2022recipe}. 
In general, these methods can be approximately divided into two parts. 
The first category mainly focuses on performing Graph Neural Networks (GNNs) on graph data.
These methods follow the convolutional pattern to define the convolution operation in the graph data, and design effective neighborhood aggregation schemes to learn node representations by fusing the node and graph topology information. 
The representative method is Graph Convolutional Network (GCN) \cite{kipf2016semi}, which learns the representation of a node in the graph by considering fusing its neighbors. 
After that, many GCN variants \cite{xu2018powerful,chen2020measuring,liu2021non,bresson2017residual} containing different neighborhood aggregation schemes have been developed. 
The second kind of method is to build graph models based on the transformer architecture.
For example, Cai and Lam \shortcite{cai2020graph} utilized the explicit relation encoding between nodes and fused them into the encoder-decoder transformer network for effective graph-to-sequence learning. 
Graphormer \cite{ying2021graphormer} established state-of-the-art performance on graph-level prediction tasks by transforming the structure and edge features of the graph into attention biases.

\begin{figure}[!t]
    \centering
    \includegraphics[width=0.99\linewidth]{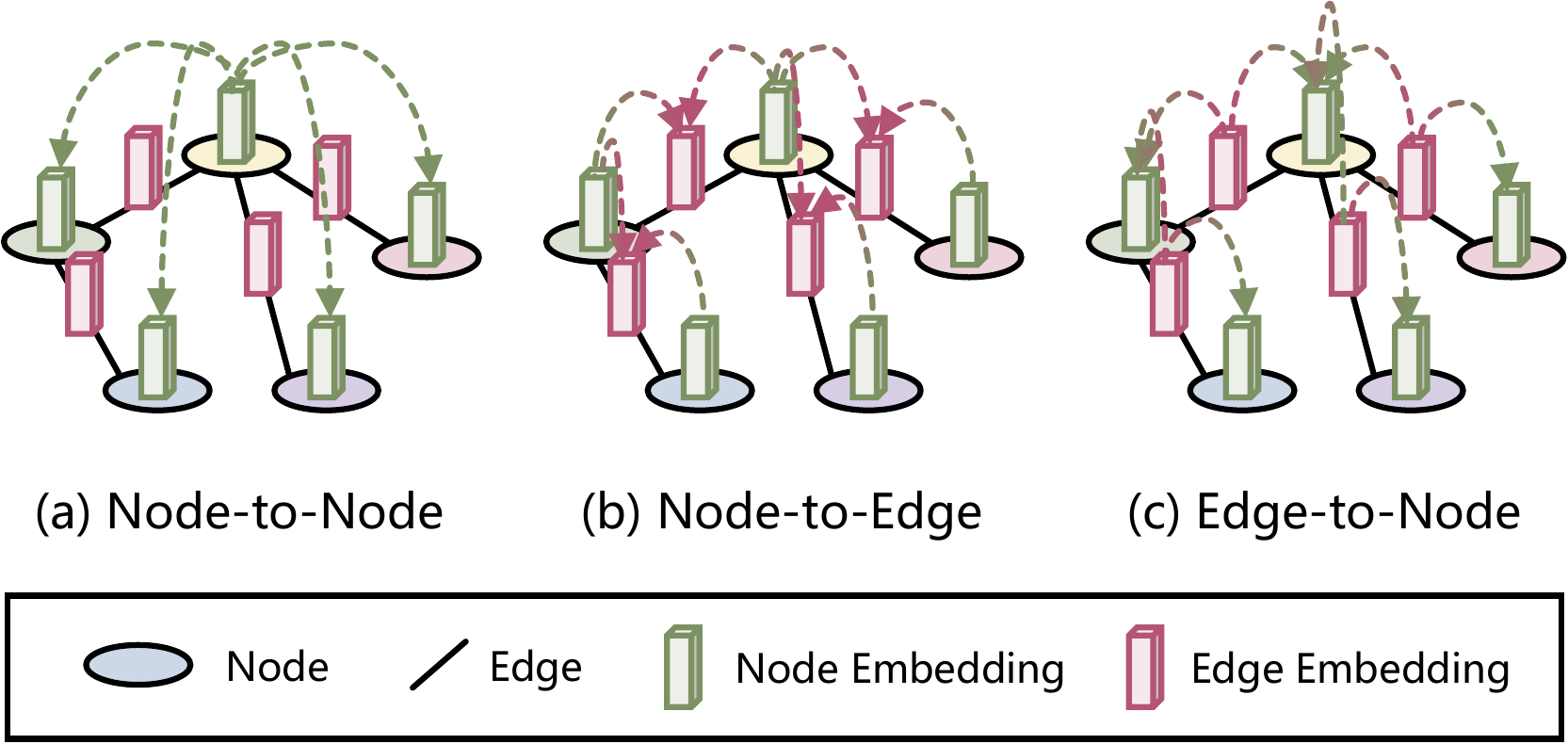}
    \caption{Illustration of the three ways for graph information propagation.
    Circles and black lines indicate nodes and edges, and green and pink cubes represent node embeddings and edge embeddings. 
    Our GPTrans achieves better graph representation learning by explicitly constructing three ways for information propagation in the proposed Graph Propagation Attention (GPA) module, including (a) node-to-node, (b) node-to-edge, and (c) edge-to-node.
    }
    \label{fig:graph_propagation}
\end{figure}

Although recent transformer-based methods report promising performance for graph representation learning, they still suffer the following problems. 
(1) \emph{Not explicitly employ the relationship among nodes and edges in the graph data.} 
Recent transformer-based methods \cite{cai2020graph,ying2021graphormer} only simply fuse nodes and edges information by using positional encodings. However, due to the complexity of graph structure, how to fully employ the relationship among nodes and edges for graph representation learning remains to be studied.  
(2) \emph{Inefficient dual-FFN structure in the transformer block.} 
Recent works resort to the dual-path structure in the transformer block to incorporate the edge information. For instance, the Edge-augmented Graph Transformer (EGT) \cite{hussain2021egt} adopted dual feed-forward networks (FFN) in the transformer block to update the edge embeddings, and let the structural information evolve from layer to layer. However, this paradigm learns the information of edges and nodes separately, which introduces more calculations and easily leads to the low efficiency of the model.

To overcome these issues, we propose an efficient and powerful transformer architecture for graph learning, termed \textbf{G}raph \textbf{P}ropagation \textbf{Trans}former (GPTrans). 
A key design element of GPTrans is its Graph Propagation Attention (GPA) module. 
As illustrated in Figure~\ref{fig:graph_propagation}, the GPA module propagates the information among the node embeddings and edge embeddings of the preceding layer by modeling three connections, \ie node-to-node, node-to-edge, and edge-to-node, which significantly enhances modeling capability (see Table~\ref{tab:pcqm4m}).
This design benefits us no longer the need to maintain an FFN module specifically for edge embeddings, bringing higher efficiency than previous dual-FFN methods.

The contributions of our work are as follows:

$(1)$ We propose an effective Graph Propagation Transformer (GPTrans), which can better model the relationship among nodes and edges and represent the graph.

$(2)$ We introduce a novel attention mechanism in the transformer blocks, which explicitly passes the information among nodes and edges in three ways. 
These relationships play a critical role in graph representation learning.

$(3)$ Extensive experiments show that the proposed GPTrans model outperforms many state-of-the-art transformer-based methods on benchmark datasets with better performance.

\section{Related Works}

\subsection{Transformer}
The past few years have seen many transformer-based models designed for various language \cite{vaswani2017attention,radford2019language,brown2020language} and vision tasks \cite{parmar2018image,liu2021swin}. 
For example, in the field of vision, Dosovitskiy \etal\shortcite{dosovitskiy2020image} presented the Vision Transformer (ViT), which decomposed an image into a sequence of patches and captured their mutual relationships.
However, training ViT on large-scale datasets can be computationally expensive. To address this issue, DeiT \cite{touvron2021training} proposed an efficient training strategy that enabled ViT to deliver exceptional performance even when trained on smaller datasets. Nevertheless, the complexity and performance of ViT remain challenging. To overcome these limitations, researchers further proposed many well-designed models \cite{liu2021swin,wang2021pyramid,wang2022internimage,chen2022vision,ji2023ddp,chen2022internvideo,wang2022internvideo}.

Recently, the self-attention mechanism and transformer architecture have been gradually introduced into the graph representation learning tasks, such as graph-level prediction \cite{ying2021graphormer,hussain2021egt}, producing competitive performance compared to the traditional GNN models.
The early self-attention-based GNNs focused on adopting the attention mechanism to a local neighborhood of each node in a graph, or directly on the whole graph. 
For example, Graph Attention Network (GAT)~\cite{velivckovic2017graph} and Graph Transformer (GT)~\cite{dwivedi2020generalization} utilized self-attention mechanisms as local constraints for the local neighborhood of each node. 
In contrast to employing local self-attention for graph learning, Graph-BERT~\cite{zhang2020graph} introduced the global self-attention mechanism in a revised transformer network to predict one masked node in a sampled subgraph. 

In addition, several works have attempted to use the transformer architecture to tackle graph-related tasks directly. 
Two notable examples are \cite{cai2020graph} and Graphormer \cite{ying2021graphormer}.
The former method adopted explicit relation encoding between nodes and integrated them into the encoder-decoder transformer network, to enable graph-to-sequence learning. 
The latter mainly regarded the structure and edges of the graph as the attention biases, which were incorporated into the transformer block. With the help of these attention biases, Graphormer achieved leading performance on graph-level prediction tasks (\eg classification and regression on molecular graphs).

\begin{figure*}[tbp]
    \centering
    \includegraphics[width=0.95\linewidth]{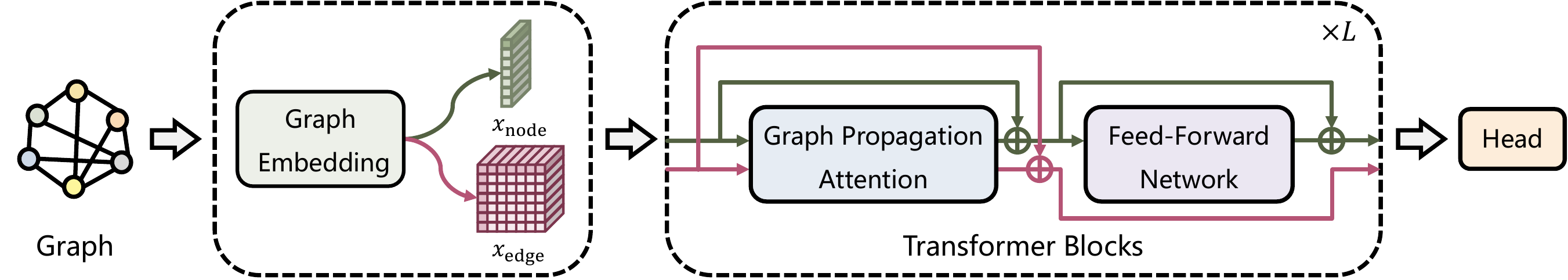}
    \caption{
        Overall architecture of GPTrans. It contains a graph embedding layer, $L$ transformer blocks, and a head.
        The graph embedding layer transforms the graph data into node embeddings $x_{\rm node}$ and edge embeddings $x_{\rm edge}$, as the input of the transformer blocks.
        Each transformer block comprises a Graph Propagation Attention (GPA) and a Feed-Forward Network (FFN).
        It is worth noting that we no longer need to maintain an FFN module specifically for edge embeddings due to the proposed GPA module, which improves the efficiency of our method.
        Finally, a head of two fully-connected layers is employed on the output embeddings for various graph tasks.
        }
    \label{fig:overall_architecture}
\end{figure*}

\subsection{Graph Convolutional Network}
Graph Convolutional Network (GCN) is a kind of deep neural network that extends the CNN from grid data (\eg image and video) to graph-structured data. 
Generally speaking, GCN methods can be approximately divided into two types: spectral-based methods~\cite{bruna2013spectral,defferrard2016convolutional,henaff2015deep,kipf2016semi} and non-spectral methods~\cite{chen2018fastgcn,gilmer2017neural,scarselli2008graph,velivckovic2017graph}. 

Spectral GCN methods are designed under the theory of spectral graphs. 
For instance, Spectral GCN~\cite{bruna2013spectral} resorted to the Fourier basis of a graph to conduct convolution operation in the spectral domain, which is the first work on spectral graph CNNs. 
Based on~\cite{bruna2013spectral}, Defferrard \etal\shortcite{defferrard2016convolutional} designed a strict control over the local support of filters and avoided an explicit use of the Graph Fourier basis in the convolution, which achieved better accuracy. Kipf and Welling \shortcite{kipf2016semi} adopted the first-order approximation of the spectral graph convolution to simplify commonly used GNN.

On the other hand, non-spectral methods directly define convolution operations on the graph data. GraphSage~\cite{hamilton2017inductive} proposed learnable aggregator functions in the network to fuse neighbors' information for effective graph representation learning. 
In GAT~\cite{velivckovic2017graph}, different weight matrices are used for nodes with different degrees for graph representation learning. 
In addition, another line of GCN methods is mainly designed for specific graph-level tasks. 
For example, some techniques such as subsampling~\cite{chen2017stochastic} and inductive representation for a large graph~\cite{hamilton2017inductive} have been introduced for better graph representative learning.

\section{GPTrans}

\subsection{Overall Architecture}

An overview of the proposed GPTrans framework is depicted in Figure~\ref{fig:overall_architecture}.
Specifically, it takes a graph $G = (V, E)$ as input, in which nodes $V=\{v_1, v_2, \dots, v_n\}$, $E$ indicates edges between nodes, and $n$ means the number of nodes.
The pipeline of GPTrans can be divided into three parts: graph embedding, transformer blocks, and prediction head.

In the graph embedding layer, for each given graph $G$, we follow \cite{ying2021graphormer,hussain2021egt} to add a virtual node $[v_0]$ into $V$, to aggregate the information of the entire graph. 
Thus, the newly-generated node set with the virtual node is represented as $V'=\{[v_0], v_1, v_2, \dots, v_n\}$, and the number of nodes is updated to $|V'|=1+n$.
For more adequate information propagation across the whole graph, each node and edge is treated as a token.
In detail, we transform the input nodes into a sequence of node embeddings $x_{\rm node}\in \mathbb{R}^{(1+n)\times d_1}$, and encode the input edges into a tensor of edge embeddings $x_{\rm edge}\in\mathbb{R}^{(1+n)\times(1+n)\times d_2}$.

Then, $L$ transformer blocks with our re-designed self-attention operation (\ie Graph Propagation Attention) are applied to node embeddings and edge embeddings. 
Both these embeddings are fed throughout all transformer blocks.
After that, GPTrans generates the representation of each node and edge, in which the output embedding of the virtual node takes along the representation of the whole graph. 

Finally, the head of our GPTrans is composed of two fully-connected (FC) layers.
For graph-level tasks, we employ it on top of the output embedding of the virtual node. 
For node-level or edge(link)-level tasks, we apply it to the output node embeddings or edge embeddings.
In summary, benefiting from the proposed novel Graph Propagation Attention, our GPTrans can better support various graph tasks with only a little additional computational cost compared to the previous method Graphormer~\cite{ying2021graphormer}.

\subsection{Graph Embedding}
The role of the graph embedding layer is to transform the graph data as the input of transformer blocks.
As we know, in addition to the nodes, edges also have rich structural information in many types of graphs, \eg molecular graphs~\cite{hu2021ogb} and social graphs~\cite{huang2022dgraph}.
Therefore, we encode both nodes and edges into embeddings to fully utilize the structure of the input graph.

% how to encode nodes
For nodes in the graph, we transform each node into node embedding. Specifically, we follow \cite{ying2021graphormer} to exploit the node attributes and the degree information, 
and add a virtual node $[v_0]$ into the graph to collect and propagate graph-level features.
Without loss of generality, taking a directed graph as an example, its node embeddings $x_{\rm node}\in \mathbb{R}^{(1+n)\times d_1}$ can be expressed as:
\begin{equation}
	x_{\rm node}= x_{\rm node\_attr} + x_{\rm deg^-} + x_{\rm deg^+},
\end{equation}
where $x_{\rm node\_attr}$, $x_{\rm deg^-}$, and $x_{\rm deg^+}$ are embeddings encoded from node attributes, indegree, and outdegree statistics, respectively.
$d_1$ is the dimension of node embeddings.

% how to encode edges
For edges in the graph, we also transform them into edge embeddings to help the learning of graph representation. 
In our implementation, the edge embeddings $x_{\rm edge}\in\mathbb{R}^{(1+n)\times(1+n)\times d_2}$ are defined as:
\begin{equation}
	x_{\rm edge}= x_{\rm edge\_attr} + x_{\rm rel\_pos},
\end{equation}
where $x_{\rm edge\_attr}$ is encoded from the edge attributes, and $x_{\rm rel\_pos}$ is a relative positional encoding that embeds the spatial location of node pairs.
$d_2$ means the dimension of edge embeddings.
We adopt the encoding of the shortest path distance by default following ~\cite{ying2021graphormer}.
In other words, for the position $(i, j)$, $x_{\rm edge}^{ij}\in\mathbb{R}^{d_2}$ represents the learned structural embedding of the edge (path) between node $v_i$ and node $v_j$ in the graph $G$.

It is worth noting that, unlike Graphormer~\cite{ying2021graphormer} that encodes edge attributes and spatial position as attention biases and shares them across all blocks, we optimize the edge embeddings $x_{\rm edge}$ in each transformer block by the proposed Graph Propagation Attention. Then, the updated edge embeddings are fed into the next block.
Therefore, each block of our model could adaptively learn different ways to exploit edge features and propagate information.
This more flexible way is beneficial for graph representation learning, which we will show in later experiments.

\begin{figure}[!t]
    \centering
    \includegraphics[width=0.8\linewidth]{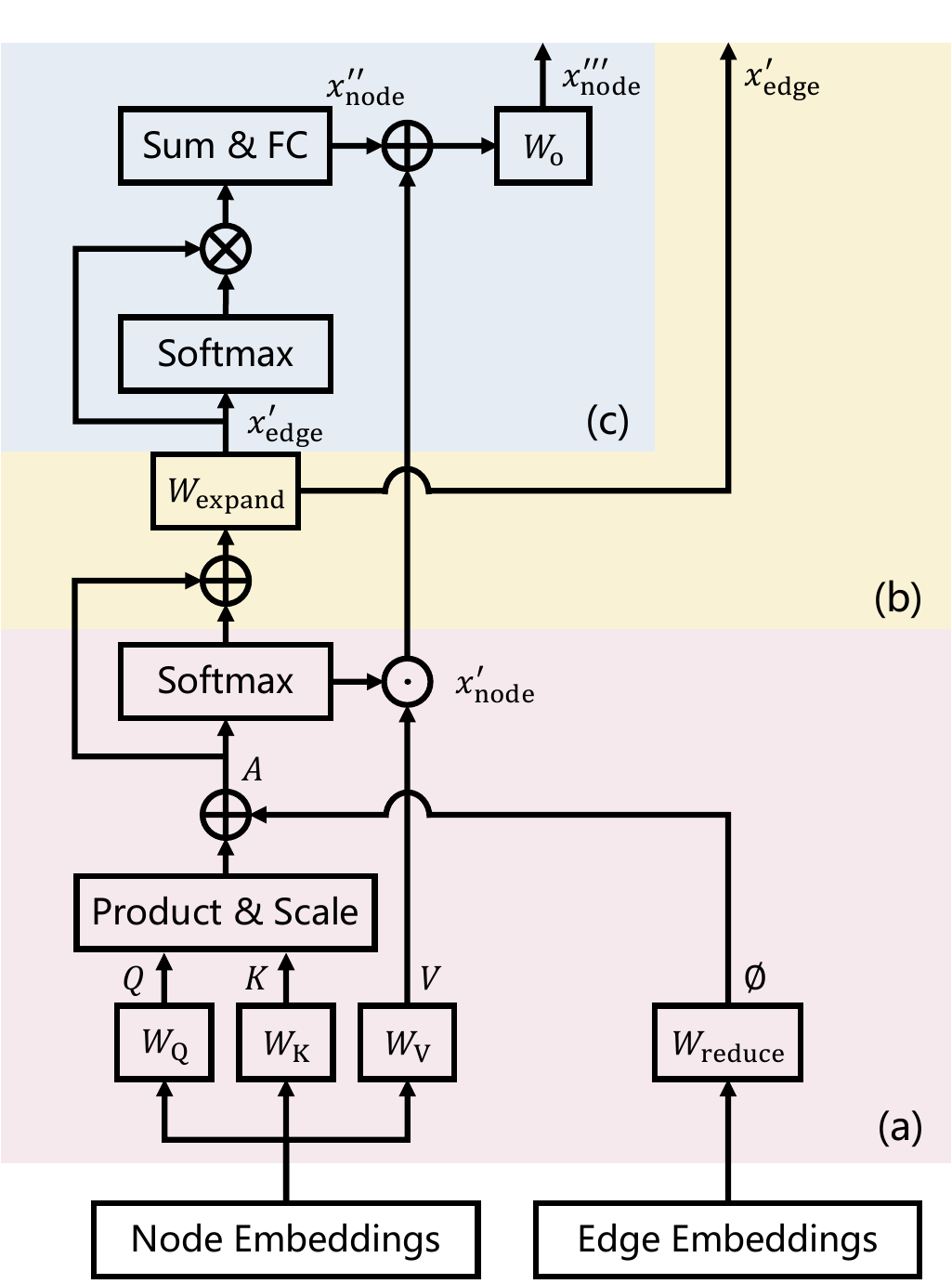}
    \caption{Illustration of Graph Propagation Attention.
    It explicitly builds three paths for information propagation among node embeddings and edge embeddings, including (a) node-to-node, (b) node-to-edge, and (c) edge-to-node.
    }
    \label{fig:gpa}
\end{figure}

\subsection{Graph Propagation Attention}

In recent years, many works~\cite{ying2021graphormer,shi2022graphormerv2,hussain2021egt} show global self-attention could serve as a flexible alternative to graph convolution and help better graph representation learning. 
However, most of them only consider part of the information propagation paths in graph, or introduce a lot of extra computational overhead to utilize edge information.
For instance, Graphormer~\cite{ying2021graphormer} only used edge features as shared bias terms to refine the attention weights of nodes. GT~\cite{dwivedi2020generalization} and EGT~\cite{hussain2021egt} designed dual-FFN networks to fuse edge features.
Inspired by this, we introduce Graph Propagation Attention (GPA), as an efficient replacement for vanilla self-attention in graph transformers. 
With an affordable cost, it could support three types of propagation paths, including node-to-node, node-to-edge, and edge-to-node.
For simplicity of description, we consider single-head self-attention in the following formulas.

\subsubsection{Node-to-Node} 
Following common practices \cite{ying2021graphormer,shi2022graphormerv2,hussain2021egt}, we adopt global self-attention \cite{vaswani2017attention} to perform node-to-node propagation. 
First, we use parameter matrices $W_{\rm Q}$, $W_{\rm K}$, and $W_{\rm V}\in \mathbb{R}^{d_1\times d_1}$ to project the node embeddings $x_{\rm node}$ to queries $Q$, keys $K$, and values $V$:
\begin{equation}
	Q = x_{\rm node}W_{\rm Q}, \ \ K = x_{\rm node}W_{\rm K}, \ \ V = x_{\rm node}W_{\rm V}.
\end{equation}
Unlike Graphormer~\cite{ying2021graphormer} that used shared attention biases in all blocks, we employ a parameter matrix $W_{\rm reduce}\in \mathbb{R}^{d_2\times {n_{\rm head}}}$ to predict layer-specific attention biases $\phi\in\mathbb{R}^{(1+n)\times(1+n)\times n_{\rm head}}$ from the edge embeddings $x_{\rm edge}$, which can be written as:
\begin{equation}
	\phi = x_{\rm edge}W_{\rm reduce}.
\end{equation}
Then, we add $\phi$ to the attention map of the query-key dot product and compute the output node embeddings $x'_{\rm node}$. This process can be formulated as:
\begin{equation}
	A =\frac{QK^T}{\sqrt{d_{\rm head}}}+\phi, \ \ \ \ x'_{\rm node} ={\rm softmax}(A)V,
\end{equation}
where $A\in\mathbb{R}^{(1+n)\times(1+n)\times n_{\rm head}}$ represents the output attention map, and $d_{\rm head}$ refers to the dimension of each head.
In summary, since $\phi$ are projected from higher dimensional edge embeddings by the learnable matrix, our attention map $A$ will have more flexible patterns to aggregate node features.

\subsubsection{Node-to-Edge}
To propagate node features to edges, we make some additional use of the attention map $A$. 
According to the definition of self-attention~\cite{vaswani2017attention}, attention map $A$ captures the similarity between node embeddings. 
Therefore, considering both local and global connections, we add $A$ with its softmax confidence, and expand its dimension to as same as $x_{\rm edge}$ through the learnable matrix $W_{\rm expand}\in\mathbb{R}^{n_{\rm head}\times d_2}$. 
This operation is designed to perform explicit high-order spatial interactions, which can be written as follows:
\begin{equation}
	x'_{\rm edge} =(A+{\rm softmax}(A))W_{\rm expand}.
	\label{hat_x_edge}
\end{equation}
In this way, we achieve node-to-edge propagation without relying on an additional FFN module like GT \cite{dwivedi2020generalization} and EGT \cite{hussain2021egt}.

\subsubsection{Edge-to-Node}
In this part, we delve into the following question: How to generate dynamic weights for edge embeddings $x_{\rm edge}$ and fuse them into node embeddings $x_{\rm node}$? 
Due to the computational efficiency, we do not additionally perform attention operation, but directly apply the softmax function to the just generated $x'_{\rm edge}\in\mathbb{R}^{(1+n)\times(1+n)\times d_2}$ in Eqn.~\ref{hat_x_edge} and calculate element-wise product with itself:
\begin{equation}
	x''_{\rm node} = {\rm FC}({\rm sum}(x'_{\rm edge}\cdot{\rm softmax}(x'_{\rm edge}), {\rm dim}=1)),
\end{equation}
in which the fully-connected (FC) layer is used to align the dimension of edge embeddings and node embeddings.
This process again explicitly introduces high-order spatial interactions.
Finally, we add these two types of node embeddings, and employ a learnable matrix $W_{\rm O}\in\mathbb{R}^{d_1\times d_1}$ to fuse them. Then we have the updated node embeddings:
\begin{equation}
	x'''_{\rm node} = (x'_{\rm node}+x''_{\rm node}) W_{\rm O}.
\end{equation}

\subsubsection{GPA in Transformer Blocks}
Equipped with our proposed GPA module, the block of our GPTrans can be calculated as follows:
\begin{equation}
	\hat{x}^{l}_{\rm node}, x^{l}_{\rm edge} \mathrel{{+}{=}} {\rm GPA}({\rm LN}(x^{l-1}_{\rm node}), x^{l-1}_{\rm edge}),
\end{equation}
\begin{equation}
	x^{l}_{\rm node} = {\rm FFN}({\rm LN}(\hat{x}^{l}_{\rm node}))+\hat{x}^{l}_{\rm node},
\end{equation}
where $\rm LN(\cdot)$ means layer normalization~\cite{ba2016layer}. 
$\hat{x}^{l}_{\rm node}$ and $x^{l}_{\rm edge}$ denote the output node embeddings and edge embeddings of the GPA
module for block $l$.
And $x^{l}_{\rm node}$ represents the output node embeddings of the FFN module.
Overall, our GPA module effectively extends the ability of our GPTrans to various graph tasks, but only introduces a small amount of extra overhead compared with previous methods \cite{hussain2021egt,ying2021graphormer}.

\subsection{Architecture Configurations}
\label{sec:architecture}

\begin{table}[t]
    \small
    \centering
    \setlength{\tabcolsep}{0.75mm}{
    \begin{tabular}{l|c|cc|cc}
    \toprule 
    % \cmidrule{3-4}
     & & \multicolumn{2}{c|}{\textbf{PCQM4M}$\downarrow$ }
     & \multicolumn{2}{c}{\textbf{PCQM4Mv2}$\downarrow$} \\
    \textbf{Model}
        & \textbf{\#Param} 
        & \textbf{Validate}
        & \textbf{Test}
        & \textbf{Validate}
        & \textbf{Test-dev}\\
    \midrule
    \multicolumn{6}{l}{\emph{Non-transformer-based Methods}} \\
    GCN 
        & 2.0M
        & 0.1684
        & 0.1838
        & 0.1379
        & 0.1398 \\
    GIN
        & 3.8M
        & 0.1536
        & 0.1678
        & 0.1195
        & 0.1218 \\
    GCN-VN 
        & 4.9M
        & 0.1510
        & 0.1579
        & 0.1153
        & 0.1152\\
    GIN-VN 
        & 6.7M
        & 0.1396
        & 0.1487
        & 0.1083
        & 0.1084\\
    GINE-VN
        & 13.2M 
        & 0.1430
        & $-$
        & $-$
        & $-$\\
    DeeperGCN-VN
        & 25.5M 
        & 0.1398
        & $-$
        & $-$
        & $-$\\
    \midrule
    \multicolumn{6}{l}{\emph{Transformer-based Methods}} \\
    GPS-Small
        & 6.2M
        & $-$
        & $-$
        & 0.0938
        & $-$\\
    \rowcolor{gray!20}
    GPTrans-T (ours)
        & 6.6M
        & \textbf{0.1179}
        & $-$
        & \textbf{0.0833}
        & $-$\\
    \midrule
    Graphormer-S
        & 12.5M 
        & 0.1264
        & $-$
        & 0.0910
        & $-$\\
    EGT-Small
        & 11.5M 
        & 0.1260
        & $-$
        & 0.0899 
        & $-$\\
    GPS-Medium
        & 19.4M 
        & $-$
        & $-$
        & 0.0858
        & $-$\\
    \rowcolor{gray!20}
    GPTrans-S (ours)
        & 13.6M 
        & \textbf{0.1162}
        & $-$
        & \textbf{0.0823}
        & $-$\\
    \midrule
    TokenGT
        & 48.5M
        & $-$
        & $-$
        & 0.0910
        & $-$ \\
    Graphormer-B
        & 47.1M 
        & 0.1234 
        & $-$
        & 0.0906 
        & $-$\\
    GRPE-Standard
        & 46.2M
        & 0.1225
        & $-$
        & 0.0890
        & 0.0898\\
    EGT-Medium
        & 47.4M 
        & 0.1224 
        & $-$
        & 0.0881 
        & $-$\\
    \rowcolor{gray!20}
    GPTrans-B (ours)
        & 45.7M 
        & \textbf{0.1153}
        & $-$
        & \textbf{0.0813}
        & $-$\\
    \midrule
    GT-Wide
        & 83.2M 
        & 0.1408
        & $-$
        & $-$ 
        & $-$\\
    GraphormerV2-L
        & 159.3M 
        & 0.1228
        & $-$
        & 0.0883 
        & $-$\\
    EGT-Large
        & 89.3M 
        & $-$
        & $-$
        & 0.0869 
        & 0.0872 \\
    EGT-Larger
        & 110.8M 
        & $-$
        & $-$
        & 0.0859 
        & $-$ \\
    GRPE-Large
        & 118.3M 
        & $-$
        & $-$
        & 0.0867 
        & 0.0876\\
    GPS-Deep
        & 138.1M 
        & $-$
        & $-$
        & 0.0852
        & 0.0862\\
    \rowcolor{gray!20}
    GPTrans-L (ours)
        & 86.0M 
        & \textbf{0.1151}
        & $-$
        & \textbf{0.0809}
        & \textbf{0.0821}\\
    \bottomrule
\end{tabular}
    }
    \caption{
	Results on PCQM4M and PCQM4Mv2.
	The metric is the Mean Absolute Error (MAE), and the lower the better.
	“$-$” denotes results are not available since the labels of test and test-dev sets are not public. 
 Highlighted are the \textbf{best} results for each model size.
    }   
    \label{tab:pcqm4m}
\end{table}

We build five variants of the proposed model with different model sizes, namely GPTrans-Nano, Tiny, Small, Base, and Large.
Note that the number of parameters of our GPTrans is similar to Graphormer~\cite{ying2021graphormer} and EGT~\cite{hussain2021egt}.
The dimension of each head is set to 10 for our nano model, and 32 for others. 
Following common practices, the expansion ratio of the FFN module is $\alpha = 1$ for all model variants.
The architecture hyper-parameters of these five models are as follows:
\begin{itemize}
    \item GPTrans-Nano: $d_1=80$, $d_2=40$, layer number = $12$
    \item GPTrans-Tiny: $d_1=256$, $d_2=32$, layer number = $12$
    \item GPTrans-Small: $d_1=384$, $d_2=48$, layer number = $12$
    \item GPTrans-Base: $d_1=608$, $d_2=76$, layer number = $18$
    \item GPTrans-Large: $d_1=736$, $d_2=92$, layer number = $24$
\end{itemize}
The model size and performance of the model variants on the large-scale PCQM4M and PCQM4Mv2 benchmarks~\cite{hu2021ogb} are listed in Table~\ref{tab:pcqm4m}, and the analysis of model efficiency is provided in Table~\ref{tab:efficiency}.
More detailed model configurations are presented in the appendix.

\begin{table}[t]
	\centering
	\small
	\setlength{\tabcolsep}{0.95mm}{
    \begin{tabular}{l|cc}
    \toprule
    \textbf{Model} 
        & \textbf{\#Param} 
        & \textbf{Test AP(\%)}$\uparrow$  \\
    \midrule 
    \multicolumn{3}{l}{\emph{Non-transformer-based Methods}} \\
    DeeperGCN-VN-FLAG~\cite{li2020deepergcn}
        & 5.6M
        & 28.42 $\pm$ 0.43  \\
    PNA~\cite{corso2020principal}
        & 6.5M 
        & 28.38 $\pm$ 0.35   \\
    DGN~\cite{beaini2021directional}
        & 6.7M 
        & 28.85 $\pm$ 0.30  \\
    GINE-VN \cite{brossard2020graph}
        & 6.1M 
        & 29.17 $\pm$ 0.15  \\
    PHC-GNN~\cite{le2021parameterized}
        & 1.7M 
        & 29.47 $\pm$ 0.26  \\
    GIN-VN$^\dagger$~\cite{xu2018powerful}
        & 3.4M 
        & 29.02 $\pm$ 0.17  \\
    \midrule
    \multicolumn{3}{l}{\emph{Transformer-based Methods}} \\
    GRPE-Standard$^\dagger$~\cite{park2022grpe} & 46.2M & 30.77 $\pm$ 0.07 \\
    \rowcolor{gray!20}
    GPTrans-B$^\dagger$ (ours) & 45.7M & \textbf{31.15 $\pm$ 0.16}\\
    \midrule
    GRPE-Large$^\dagger$~\cite{park2022grpe} & 118.3M & 31.50 $\pm$ 0.10 \\
    Graphormer-L$^\dagger$~\cite{ying2021graphormer}
        & 119.5M 
        & 31.39 $\pm$ 0.32 \\
    EGT-Larger$^\dagger$ \cite{hussain2021egt}
        & 110.8M 
        & 29.61 $\pm$ 0.24  \\
    \rowcolor{gray!20} 
    GPTrans-L$^\dagger$ (ours) & 86.0M & \textbf{32.43 $\pm$ 0.22} \\
    \bottomrule
\end{tabular}}
    \caption{Results on MolPCBA. $^\dagger$ indicates the model is pre-trained on PCQM4M or PCQM4Mv2.
The higher the better. 
 % Highlighted are the top {\color{magenta}{\textbf{first}}} and {\color{violet}{\textbf{second}}} results for each model size.
 Highlighted are the \textbf{best} results for each model size.
 }
\label{tab:molpcba}
\end{table}

\begin{table}[t]

	\centering
	\small
	\setlength{\tabcolsep}{1.45mm}{
    
\begin{tabular}{l|cc}
    \toprule
    \textbf{Model} 
        & \textbf{\#Param} 
        & \textbf{Test AUC(\%)}$\uparrow$ \\
    \midrule
    \multicolumn{3}{l}{\emph{Non-transformer-based Methods}} \\
    DeeperGCN-FLAG~\cite{li2020deepergcn}
        & 532K 
        & 79.42 $\pm$ 1.20 \\
    PNA~\cite{corso2020principal}
        & 326K 
        & 79.05 $\pm$ 1.32 \\
    DGN~\cite{beaini2021directional}
        & 110K 
        & 79.70 $\pm$ 0.97 \\
    PHC-GNN~\cite{le2021parameterized} 
        & 114K 
        & 79.34 $\pm$ 1.16 \\
    GIN-VN$^\dagger$~\cite{xu2018powerful}
        & 3.3M 
        & 77.80 $\pm$ 1.82 \\
    \midrule
    \multicolumn{3}{l}{\emph{Transformer-based Methods}} \\
    Graphormer-B$^\dagger$~\cite{ying2021graphormer}
        & 47.0M 
        & 80.51 $\pm$ 0.53 \\
    EGT-Larger$^\dagger$ \cite{hussain2021egt}
        & 110.8M 
        & 80.60 $\pm$ 0.65 \\
    GRPE-Standard$^\dagger$~\cite{park2022grpe}
        & 46.2M 
        & \textbf{81.39 $\pm$ 0.49} \\
    \rowcolor{gray!20}
    GPTrans-B$^\dagger$ (ours) & 45.7M & 81.26 $\pm$ 0.32 \\
    % \midrule

    % \rowcolor{gray!20}
    % GPTrans-L$^\dagger$ (ours)& 86.0M & \textbf{81.50 $\pm$ 0.59} \\
    \bottomrule
\end{tabular}}
    \caption{Results on MolHIV. $^\dagger$ indicates the model is pre-trained on PCQM4M or PCQM4Mv2.
    The higher the better.
 % Highlighted are the top {\color{magenta}{\textbf{first}}} and {\color{violet}{\textbf{second}}} results for each model scale.
    Highlighted are the \textbf{best} results.
}
	\label{tab:molhiv}
\end{table}

\section{Experiments}

\begin{table*}[t]
    \centering
    \small
    \setlength{\tabcolsep}{3.15mm}{
    \begin{tabular}{l|c|c|c|c|c}
    \toprule
        & 
        & \textbf{ZINC}
        & \textbf{PATTERN} 
        & \textbf{CLUSTER} 
        & \textbf{TSP}\\
    \textbf{Model} 
    & \textbf{\#Param}
    & \textbf{Test MAE}$\downarrow$ 
    & \textbf{Accuracy(\%)}$\uparrow$ 
    & \textbf{Accuracy(\%)}$\uparrow$
    & \textbf{F1-Score}$\uparrow$\\
    \midrule
    \multicolumn{3}{l}{\emph{Non-transformer-based Methods}} \\
    GCN~\cite{kipf2016semi}
        &  505K
        &  0.367 $\pm$ 0.011
        &  71.892 $\pm$ 0.334
        &  68.498 $\pm$ 0.976
        &  $-$ \\
    GraphSage~\cite{hamilton2017inductive}
        &  505K
        &  0.398 $\pm$ 0.002
        &  50.492 $\pm$ 0.001
        &  63.884 $\pm$ 0.110
        &  $-$\\
    GIN~\cite{xu2018powerful}
        &  510K
        &  0.526 $\pm$ 0.051
        &  85.387 $\pm$ 0.136
        &  64.716 $\pm$ 1.553
        & $-$ \\
    GAT~\cite{velivckovic2017graph}
        &  531K
        &  0.384 $\pm$ 0.007
        &  78.271 $\pm$ 0.186
        &  70.587 $\pm$ 0.447
        &  $-$ \\
    GatedGCN~\cite{bresson2017residual} 
    % (Bresson et al. 2017)
        &  505K
        &  0.214 $\pm$ 0.013
        &  86.508 $\pm$ 0.085 
        &  76.082 $\pm$ 0.196 
        &  \textbf{\color{violet}{0.838 $\pm$ 0.002}} \\
    PNA~\cite{corso2020principal}
        & 387K
        & 0.142 $\pm$ 0.010
        & $-$ 
        & $-$
        & $-$\\
    \midrule
    \multicolumn{3}{l}{\emph{Transformer-based Methods}} \\
    GT~\cite{dwivedi2020generalization} 
        &  589K
        &  0.226 $\pm$ 0.014
        &  84.808 $\pm$ 0.068
        &  73.169 $\pm$ 0.622 
        &  $-$\\
    SAN~\cite{kreuzer2021rethinking}
        & 509K
        & 0.139 $\pm$ 0.006 
        & 86.581 $\pm$ 0.037
        & 76.691 $\pm$ 0.650
        & $-$\\
    Graphormer-Slim~\cite{ying2021graphormer}
        &  489K
        &  0.122 $\pm$ 0.006
        &  86.650 $\pm$ 0.033 
        &  74.660 $\pm$ 0.236 
        &  0.698  $\pm$ 0.007\\
    % GRPE-Small~\cite{park2022grpe}
    %     &  489K
    %     &  0.094 $\pm$ 0.002
    %     &  $-$
    %     &  $-$
    %     &  $-$\\
    EGT~\cite{hussain2021egt}
        &  500K
        &  0.108 $\pm$ 0.009
        &  \textbf{\color{magenta}{86.821 $\pm$ 0.020}}
        &  \textbf{\color{magenta}{79.232 $\pm$ 0.348}}
        &  \textbf{\color{magenta}{0.853  $\pm$ 0.001}} \\
    GPS~\cite{rampavsek2022recipe}
        &  424K
        &  \textbf{\color{magenta}{0.070 $\pm$ 0.004}}
        &  86.685 $\pm$ 0.059
        &  78.016 $\pm$ 0.180
        &  $-$\\
    \rowcolor{gray!20}
    GPTrans-Nano (ours) 
        & 554K
        & \textbf{\color{violet}{0.077 $\pm$ 0.009}}
        & \textbf{\color{violet}{86.731 $\pm$ 0.085}}
        & \textbf{\color{violet}{78.069 $\pm$ 0.154}}
        & 0.832 $\pm$ 0.004 \\
    \bottomrule
\end{tabular}}
    \caption{Results on four benchmarking datasets from \protect\cite{dwivedi2020benchmarking}, including graph regression (ZINC), node classification (PATTERN and CLUSTER), and edge classification (TSP) tasks. The arrow next to the metric means higher or lower is better. ``$-$'' denotes the results are not available. Highlighted are the top \textcolor{magenta}{\textbf{first}} and \textcolor{violet}{\textbf{second}} results.}
	\label{tab:benchmarking}
\end{table*}

\subsection{Graph-Level Tasks}
\label{sec:graph_level}
\subsubsection{Datasets}
We verify the following graph-level tasks:

\noindent (1) PCQM4M~\cite{hu2021ogb} is a quantum chemistry dataset that includes 3.8 million molecular graphs and a total of 53 million nodes. The task is to regress a DFT-calculated quantum chemical property, \eg HOMO-LUMO energy gap. 

\noindent (2) PCQM4Mv2~\cite{hu2021ogb} is an updated version of PCQM4M, in which the number of molecules slightly decreased, and some of the graphs are revised.

\noindent (3) MolHIV~\cite{hu2020open} is a small-scale molecular property prediction dataset. It has $ 41,127$ graphs with a total of $1,048,738$ nodes and $1,130,993$ edges. 

\noindent (4) MolPCBA~\cite{hu2020open} is another property prediction dataset, which is larger than MolHIV. It contains $437,929$ graphs with $11,386,154$ nodes and $12,305,805$ edges.

\noindent (5) ZINC~\cite{dwivedi2020benchmarking} is a popular real-world molecular dataset for graph property regression.
It has $10,000$ train, $1,000$ validation, and $1,000$ test graphs.

\subsubsection{Settings}
For the large-scale PCQM4M and PCQM4Mv2 datasets, we use AdamW~\cite{loshchilov2018decoupled} with an initial learning rate of 1e-3 as the optimizer. 
Following common practice, we adopt a cosine decay learning rate scheduler with a 20-epoch warmup.
All models are trained for 300 epochs with a total batch size of 1024.
When fine-tuning the MolHIV and MolPCBA datasets, we load the PCQM4Mv2 pre-trained weights as initialization.
For the ZINC dataset, we train our GPTrans-Nano model from scratch. 
More detailed training strategies are provided in the appendix.

\subsubsection{Results}
First, we benchmark our GPTrans method on PCQM4M and PCQM4Mv2, two datasets from OGB large-scale challenge \cite{hu2021ogb}.
We mainly compare our GPTrans against a set of representative transformer-based methods, including GT \cite{dwivedi2020generalization}, Graphormer \cite{ying2021graphormer}, GRPE \cite{park2022grpe}, EGT \cite{hussain2021egt}, GPS \cite{rampavsek2022recipe}, and TokenGT \cite{kim2022tokengt}.
As reported in Table~\ref{tab:pcqm4m}, our method yields the state-of-the-art validate MAE score on both datasets across different model complexities.

Further, we take the PCQM4Mv2 pre-trained weights as the initialization and fine-tune our models on the OGB molecular datasets MolPCBA and MolHIV, to verify the transfer learning capability of GPTrans. 
All experiments are performed five times with different random seeds, and we report the mean and standard deviation of the results.
From Table~\ref{tab:molpcba} and~\ref{tab:molhiv}, we can see that GPTrans outperforms many strong counterparts, such as GRPE~\cite{park2022grpe}, EGT~\cite{hussain2021egt}, and Graphormer~\cite{ying2021graphormer}.

Moreover, we follow previous methods~\cite{park2022grpe,ying2021graphormer} to train the GPTrans-Nano model with about 500K parameters on the ZINC subset from scratch. 
As demonstrated in Table~\ref{tab:benchmarking}, our model achieves a promising test MAE of 0.077 $\pm$ 0.009, bringing 36.9\% relative MAE decline compared to Graphormer~\cite{ying2021graphormer}. 
The above inspiring results show that the proposed GPTrans performs well on graph-level tasks.

\subsection{Node-Level Tasks}

\subsubsection{Datasets}
PATTERN and CLUSTER~\cite{dwivedi2020benchmarking} are both synthetic datasets for node classification. Specifically, PATTERN has $10,000$ training, $2,000$ validation, and $2,000$ test graphs, and CLUSTER contains $10,000$ training, $1,000$ validation, and $1,000$ test graphs.

\subsubsection{Settings}
For the PATTERN and CLUSTER datasets, we train our GPTrans-Nano up to 1000 epochs with a batch size of 256.
We employ the AdamW~\cite{loshchilov2018decoupled} optimizer with a 20-epoch warmup.
The learning rate is initialized to 5e-4, and is declined by a cosine scheduler. More training details can be found in the appendix.

\subsubsection{Results}
In this part, we compare our GPTrans-Nano with various GCN variants and recent graph transformers.
As shown in Table~\ref{tab:benchmarking}, our GPTrans-Nano produces the promising accuracy of 86.731 $\pm$ 0.085$\%$ and 78.069 $\pm$ 0.154$\%$ on the PATTERN and CLUSTER datasets, respectively.
These results outperform many Convolutional/Message-Passing Graph Neural Networks by large margins, showing that the proposed GPTrans can serve as an alternative to traditional GCNs for node-level tasks.
Moreover, we find that our method exceeds Graphormer \cite{ying2021graphormer} on the CLUSTER dataset by significant gaps of 3.4\% accuracy, which suggests that the three propagation ways explicitly constructed in the GPA module are also helpful for node-level tasks.

\subsection{Edge-Level Tasks}

\subsubsection{Datasets}
% TSP
TSP \cite{dwivedi2020benchmarking} is a dataset for the Traveling Salesman Problem, which is an NP-hard combinatorial optimization problem. The problem is reduced to a binary edge classification task, where edges in the TSP tour have positive labels. TSP dataset has $10,000$ training, $1,000$ validation, and $1,000$ test graphs.

\begin{table}[t]
    \small
    \centering
    \setlength{\tabcolsep}{1.45mm}
    \begin{tabular}{l|cc|c}
    \toprule 
    \textbf{Model}
        & \textbf{\#Param} 
        & \textbf{FLOPs}
        & \textbf{Validate MAE}$\downarrow$ \\
    \midrule
    Baseline (Graphormer-S) & 12.5M &  0.399G      & 0.0928 \\
    + Node-to-Node          & 13.3M &      0.402G  & 0.0874 \\ 
    ++ Node-to-Edge         & 13.3M &      0.405G  & 0.0865 \\
    +++ Edge-to-Node        & 13.5M &      0.417G  & 0.0854 \\
    \midrule
    \rowcolor{gray!5}
    GPTrans-S$_{\rm wider}$  & 13.5M &   0.417G & 0.0854 \\
    \rowcolor{gray!20}
    GPTrans-S$_{\rm deeper}$ (ours)  & 13.6M & 0.472G  & \textbf{0.0835} \\
    \bottomrule
\end{tabular}
    \caption{Ablation studies of GPTrans. We build our baseline based on Graphormer-S \protect\cite{ying2021graphormer} with a shorter schedule of 100 epochs, and decline its validate MAE on PCQM4Mv2 \protect\cite{hu2021ogb} dataset from 0.0928 to 0.0854 by gradually introducing our GPA module. Moreover, we find that the deeper model outperforms the wider model with a similar number of parameters.}
    \label{tab:ablation}
\end{table}

\subsubsection{Settings}
We experiment on the TSP dataset in a similar setting to that used in the PATTERN and CLUSTER datasets.
Details are shown in the appendix.

\subsubsection{Results}
Table~\ref{tab:benchmarking} compares the edge classification performance of our GPTrans-Nano model and previously transformer-based methods on the TSP dataset. 
We observe GPTrans can outperform Graphormer \cite{ying2021graphormer} with a large margin and is comparable with EGT \cite{hussain2021egt}, showing that the proposed GPA module design is competitive when used for edge-level tasks.
By applying the GPA module, we avoid designing an inefficient dual-FFN network, which boosts the efficiency of our method. 
We will analyze the efficiency of GPTrans in detail in Section~\ref{sec:ablation}.

\subsection{Ablation Study}
\label{sec:ablation}
We conduct several ablation studies on the PCQM4Mv2~\cite{hu2021ogb} dataset, to validate the effectiveness of each key design in our GPTrans.
Due to the limited computational resources, we adopt GPTrans-S as the base model, and train it with a shorter schedule of 100 epochs.
Other settings are the same as described in Section~\ref{sec:graph_level}.

\subsubsection{Graph Propagation Attention} 
To investigate the contribution of each key design in our GPA module, we gradually extend the Graphormer baseline \cite{ying2021graphormer} to our GPTrans.
As shown in Table~\ref{tab:ablation}, the model gives the best performance when all three information propagation paths are introduced.
It is worth noting that the improvement from our node-to-node propagation is most significant, thanks to learning the attention biases for a particular layer rather than sharing them across all layers.
In summary, our proposed GPA module collectively brings a large gain to Graphormer, \ie 8.0\% relative validate MAE decline on the PCQM4Mv2 dataset.

\subsubsection{Deeper \emph{vs.} Wider} 
Here we explore the question of whether the transformers for graph representation learning should go deeper or wider.
For fair comparisons, we build a deeper but thinner model under comparable parameter numbers, by increasing the depth from 6 to 12 layers and decreasing the width from 512 to 384 dimensions.
As reported in Table~\ref{tab:ablation}, the validate MAE of the PCQM4Mv2 dataset is declined from 0.0854 to 0.0835 by the deeper model, which shows that depth is more important than width for graph transformers.
Based on this observation, we prefer to develop GPTrans with a large model depth.

\begin{table}[t]
    \small
    \centering
    \setlength{\tabcolsep}{0.75mm}{
    \begin{tabular}{l|c|cc|c}
    \toprule 
    % \cmidrule{3-4}
     & & \textbf{Train} & \textbf{Inference} & \textbf{PCQM4Mv2}  \\
    \textbf{Model}
        & \textbf{\#Param} 
        & \textbf{(min / ep.)}
        & \textbf{(graph / s)}
        & \textbf{Validate MAE}$\downarrow$\\
    \midrule
    % Graphormer-S & 12.5M &       &   & 0.0910 \\
    EGT-Small     & 11.5M & 7.6    &    10291.8    & 0.0899 \\ 
    \rowcolor{gray!20}
    GPTrans-S     & 13.6M & 5.5    &    11391.2 & \textbf{0.0823}  \\
    \midrule
    EGT-Medium    & 47.4M & 11.3    &    4840.8 & 0.0881 \\ 
    \rowcolor{gray!20}
    GPTrans-B     & 45.7M & 7.7    &    6670.6  & \textbf{0.0813}  \\
    \midrule
    EGT-Large     & 89.3M & 15.5    &   3759.4   & 0.0869 \\ 
    \rowcolor{gray!20}
    GPTrans-L     & 86.0M & 9.6   &   4193.4 & \textbf{0.0809}  \\
    \bottomrule
\end{tabular}}
    \caption{Efficiency analysis of GPTrans. 
    These experiments are conducted with PyTorch1.12 and CUDA11.3.
    Training time is measured on 8 A100 GPUs with half-precision training, and the inference throughput is tested on a single A100 GPU. 
    }
    \label{tab:efficiency}
\end{table}

\subsubsection{Efficiency Analysis}
As shown in Table~\ref{tab:efficiency}, we benchmark the training time and inference throughputs of our GPTrans and EGT \cite{hussain2021egt}.
Specifically, we employ PyTorch1.12 and CUDA11.3 to perform these experiments.
For a fair comparison, the training time of these two methods is measured using 8 Nvidia A100 GPUs with half-precision training and a total batch size of 1024.
The inference throughputs of PCQM4Mv2 models in Table \ref{tab:efficiency} are tested using an A100 GPU with a batch size of 128, where our GPTrans is slightly faster in inference than EGT under a similar number of parameters.
This preliminary study shows a good signal that the proposed GPTrans, equipped with the GPA module, could be an efficient model for graph representation learning.

\section{Conclusion}
This paper aims for graph representation learning with a Graph Propagation Transformer (GPTrans), which explores the information propagation among nodes and edges in a graph when establishing the self-attention mechanism in the transformer block. 
Especially in the GPTrans, we propose a Graph Propagation Attention (GPA) mechanism to explicitly pass the information among nodes and edges in three ways, \ie node-to-node, node-to-edge, and edge-to-node, which is essential for learning graph-structured data. Extensive comparisons with state-of-the-art methods on several benchmark datasets demonstrate the superior capability of the proposed GPTrans with better performance.

\section*{Acknowledgements}
This work is supported by the Natural Science Foundation of China under Grant 61672273 and Grant 61832008.

\section*{Contribution Statement}
Zhe Chen and Hao Tan contributed equally to this work and are listed as co-first authors.
Tong Lu served as the corresponding author for communication and correspondence.

{
\small
\bibliographystyle{named}
\bibliography{ijcai23}
}

\end{document}